\documentclass[sigconf]{aamas}

\usepackage{balance} 
\usepackage{graphicx} 
\usepackage{subfigure}
\usepackage{amsmath}

\usepackage{algorithm}
\usepackage{amsfonts}       
\usepackage{algpseudocode}  

\setcopyright{ifaamas}
\acmConference[AAMAS '24]{Proc.\@ of the 23rd International Conference
on Autonomous Agents and Multiagent Systems (AAMAS 2024)}{May 6 -- 10, 2024}
{Auckland, New Zealand}{N.~Alechina, V.~Dignum, M.~Dastani, J.S.~Sichman (eds.)}
\copyrightyear{2024}
\acmYear{2024}
\acmDOI{}
\acmPrice{}
\acmISBN{}

\acmSubmissionID{467}

\title[AAMAS-2024 Formatting Instructions]{Sample Efficient Reinforcement Learning by Automatically Learning to Compose Subtasks}

\author{Shuai Han}
\affiliation{
	\institution{Utrecht University}
	\city{Utrecht}
	\country{the Netherland}}
\email{s.han@uu.nl}

\author{Mehdi Dastani}
\affiliation{
	\institution{Utrecht University}
	\city{Utrecht}
	\country{the Netherland}}
\email{m.m.dastani@uu.nl}

\author{Shihan Wang}
\affiliation{
	\institution{Utrecht University}
	\city{Utrecht}
	\country{the Netherland}}
\email{s.wang2@uu.nl}

\begin{abstract}
Improving sample efficiency is central to Reinforcement Learning (RL), especially in environments where the rewards are sparse. Some recent approaches have proposed to specify reward functions as manually designed or learned reward structures whose integrations in the RL algorithms are claimed to significantly improve the learning efficiency. Manually designed reward structures can suffer from inaccuracy and existing automatically learning methods are often computationally intractable for complex tasks. The integration of inaccurate or partial reward structures in RL algorithms fail to learn optimal policies. In this work, we propose an RL algorithm that can automatically structure the reward function for sample efficiency, given a set of labels that signify subtasks. Given such minimal knowledge about the task, we train a high-level policy that selects optimal sub-tasks in each state together with a low-level policy that efficiently learns to complete each sub-task. We evaluate our algorithm in a variety of sparse-reward environments. The experiment results show that our approach significantly outperforms the state-of-art baselines as the difficulty of the task increases.

\end{abstract}

\keywords{Reinforcement learning, Sample efficiency, Subtasks}

\newcommand{\BibTeX}{\rm B\kern-.05em{\sc i\kern-.025em b}\kern-.08em\TeX}

\begin{document}


\pagestyle{fancy}
\fancyhead{}


\maketitle 


\section{Introduction}
\label{C:Introduction}

As a powerful technique to optimize the intelligent behaviors of agents, reinforcement learning (RL) has been applied in a variety of different domains, such as traffic signal control \cite{traffic2020, traffic2022}, chemical structure prediction \cite{chm22-1, chm22-2}, radio resource management \cite{radio-suvery, radio} and games \cite{mnih2015human}. The successful training of RL agents often relies on handcrafted reward functions based on domain knowledge, which allows agents to receive immediate reward signals. Without those handcrafted signals, the sparse rewards can result in RL algorithms suffering from low sample efficiency \cite{her, gupta2022unpacking}.

Specifying the structure for a reward function with high-level logical languages or finite state machines has shown to be beneficial for improving the sample efficiency of RL algorithms \cite{TLrl2018, TL4, rm-rs}. Recently, QRM \cite{QRM} has been proposed for RL agents to exploit the specified reward structure using a reward machine. For example, when the task for the agent is to bring the coffee to the office, the corresponding reward structure is that the agent will receive reward if it first picks up the coffee and then arrives at the office. By exposing this reward structure as an reward machine and utilizing it for training the agent, the sample efficiency of RL algorithms can be significantly improved \cite{krrrl-survey}. It should be noted that such specified reward function can also be used for other purposes such as task decomposition \cite{QRM-j}, reward shaping \cite{rm-rs} and exploration \cite{rm-exp} to make the RL algorithm more efficient.

However, for complex applications, the reward structure is not always available due to a lack of sufficient domain knowledge. As emphasized in \cite{DBLP:conf/aaai/HasanbeigJAMK21}, high-level objectives corresponding to subtasks can often be identified by passing through designated and semantically distinguishable states of the environment. Nevertheless, how those high-level subtasks structure the reward is not straightforward to be handcrafted \cite{toro2019learning}. Oftentimes, such high-level knowledge is implicit and unknown to the learning agent \cite{DBLP:conf/aips/0005GAMNT020}. In such cases, it is crucial to automatically learn the structure of reward function based on the limited domain knowledge. To solve this problem, previous work has proposed to use automata learning to infer a automaton model to describe and exploit the reward function structure \cite{DBLP:conf/aips/0005GAMNT020, DBLP:conf/aaai/HasanbeigJAMK21}. However, learning an exact automaton model from trace data is a NP-complete problem \cite{DBLP:journals/iandc/Gold78}. Although heuristic methods can be used to speed up the learning \cite{oncina1992inferring}, inferring an automaton that is representative to the reward structure relies on trace data with adequately exploration. When the exploration of agent is inadequate, the automaton model derived from the incomplete trace data could be either inaccurate or partial, which leads to the RL algorithm learning sub-optimal policies or even failing to learn.

In this paper, we propose a novel RL algorithm, which we call Automatically Learning to Compose Subtasks (ALCS), to automatically exploit a minimal domain knowledge to structure the reward function for sample efficiency. Given such minimal knowledge to signify subtasks, the basic idea of ALCS is to learn the best sequences of subtasks with respect to the environment reward. To realize this, we develop a framework with two-level hierarchy of policy learning. The low-level policy learns to take the next action toward completing the given subtask, while the high-level policy learns to specify a subtask to be achieved at each time step to the low-level policy. There are two main characteristics in the high-level policy learning. One is that the next subtask is selected based on the sequence of completed subtasks, which allows our method to be more accurate to make decisions to compose subtasks. Another characteristic is that during training the high-level subtask selection is modified based on the actually achieved subtasks, which can ensure that the high-level policy does not miss any necessary subtasks in the process of reinforcing subtask selection based on environmental rewards.

This paper makes the following contributions: 1) we design a two-level RL learning framework to accurately and automatically discover and exploit reward structures through composition of subtasks, which improves sample efficiency in sparse-reward environments, 2) we design high-level subtask modification mechanism to further improve the overall policy learning, and 3) we verify the performance of our approach on $8$ sparse-reward environments and show the interpretability of the agent by producing the composed sequences of subtasks. When the difficulty of tasks increases, our method produces a significant improvement over the previous most sample-efficient methods given the same minimal domain knowledge on such domain.

\section{Related Work}
\label{C:Related_Work}

Improving sample efficiency is a key challenging problem for RL algorithms. Among the plethora of work on sample efficiency \cite{krrrl-survey, guo2022survey, moerland2023model, lyu2019sdrl, jin2022creativity}, we summarize the literature in three subfields, which our approach either builds on or closely relates to.

\textbf{Specifying the structure of reward function.} Recently, using domain knowledge to specify a reward structure that can be integrated into the RL algorithm has achieved significant improvement on the learning efficiency in sparse-reward environments. There are two major ways to provide this specified structure: Temporal Logics \cite{TL1977} and Mealy Machines \cite{Mealy}. With a logic specification for task, Temporal Logics have been applied to synthesize policies \cite{TL1, TL3} and shape reward \cite{TLrl2018, TL4} for sample efficiency. By specifying reward structure with Mealy Machines, named reward machines, QRM \cite{QRM, QRM-j} is proposed to improve the sample efficiency for RL by exposing reward structure to the learning agent. The integration of reward machines in reinforcement learning has lead to a series of proposals for exploration \cite{rm-exp}, reward shaping \cite{rm-rs}, offline learning \cite{rm-offline} and multi-agent learning \cite{rm-marl-1, rm-marl-2}. In line with these works, we also assume the domain knowledge is helpful in specifying a reward structure for sample efficiency. Our method is different from these methods in terms of the problem setting and solution. In our setting, the required domain knowledge is limited to the specification of the subtasks and not their ordering structures such as in \cite{TL4, QRM-j, rm-rs}. In our method, the ordering structure can be automatically learned, rather than manually designed.

\textbf{Learning the reward structure.} When the domain knowledge minimally structures the reward function, (i.e., only specifying the possible subtasks), there are previous methods to learn the unknown reward structure for the task from trace data. LRM \cite{toro2019learning} first proposed to automatically learn a reward machine to specify the reward structure in partially observable settings. Moreover, JIRP \cite{DBLP:conf/aips/0005GAMNT020} learns a smallest reward machine with automata learning techniques \cite{neider2013regular, jeppu2020learning}. SRM \cite{corazza2022reinforcement} considers randomness in environment rewards and learns a stochastic reward machine for task structure. Another research route is to use an automaton model to structure the reward of a task, instead of using it as a reward machine. ISA \cite{furelos2020induction} learns an automaton whose transitions are subgoals given by domain knowledge. DeepSynth \cite{DBLP:conf/aaai/HasanbeigJAMK21} employs an automata synthesis method to automatically reveal the sequential structure of given high-level objectives.

Similar to the above mentioned  approaches, our method also assumes the availability of limited domain knowledge for improving sample efficiency of RL. The difference is that our approach does not learn an exact automata model to be integrated into the RL algorithm for sample efficiency. The proposals to learn an exact automata model from the trace data is NP-complete \cite{DBLP:journals/iandc/Gold78}. Besides, when the learned automaton model is inaccurate or partial for revealing complex reward structure, the integrated RL algorithm may fail to learn the optimal policy. In contrast, our approach avoids solving an NP-complete problem by using a high-level policy, which is induced directly from the real environment rewards. This allows our high-level policy to more accurately capture the reward structure.

\textbf{Goal-conditioned reinforcement learning.} Goal-conditioned reinforcement learning (GCRL) trains an agent to achieve different goals \cite{chane2021goal, liu2022goal, colas2022autotelic} by training a goal-conditioned policy. The intrinsic rewards provided by goal accelerate learning \cite{her}. The training of goal-conditioned policy can be improved with efficient augmentation, such as curiosity exploration \cite{fang2019curriculum}, contrastive representations \cite{eysenbach2022contrastive} and causal relations \cite{ding2022generalizing}. Our approach follows the general training paradigm of the GCRL summarized in \cite{colas2022autotelic}. From the perspective of this paradigm, we propose to learn to prioritize the goal selection based on the sequence of history goals. 

\textbf{Hierarchical reinforcement learning.} Hierarchical Reinforcement Learning (HRL) methods exploit temporal abstraction \cite{hrlsurvey} or spatial abstraction \cite{baram2016spatio, zadem2023goal} to improve sample efficiency of RL. When designing the two-level policy, we are inspried by HRL with temporal abstraction, such as HAM \cite{parr1997reinforcement}, MAXQ \cite{dietterich2000hierarchical} and h-DQN \cite{kulkarni2016hierarchical}. However, those methods abstract the environment as a Semi-MDP \cite{sutton1999between}, which specifies different time scales for different level of policies. Instead, our high-level and low-level policies make decisions with the same time scale, i.e., at each time step high-level and low-level policies make joint decisions to produce actions. An HRL method with a similar decision-making process to ours is Interrupting Options \cite{sutton1999between}, in whose extreme case the option is interrupted at every step. Another difference is that our high-level policy takes into account the sequence of historically completed subtasks when making decisions. Such a non-Markovian high-level policy is helpful to structure rewards in a more accurate way. While HRL approaches select subtasks based on the environment state, which leads to HRL being inappropriate for solving the problem of structuring rewards using domain knowledge.


\section{Problem Setting}
\label{C:Preliminaries}

The RL problem considers an agent interacting with an unknown environment \cite{rlbook}. Such environment can be modeled as a Markov Decision Process (MDP), $\mathcal{M} = (\mathcal{S}, \mathcal{A}, \mathcal{T}, R, \gamma)$ where $\mathcal{S}$ is a finite set of states, $\mathcal{A}$ is a finite set of actions, $\mathcal{T}: \mathcal{S} \times \mathcal{A} \times \mathcal{S} \rightarrow [0, 1]$ is a transition function, $\gamma \in [0, 1)$ is a discount factor and $R: \mathcal{S} \times \mathcal{A} \times \mathcal{S} \rightarrow \mathbb{R}$ is a reward function. An agent employs a deterministic policy $\pi:\mathcal{S} \rightarrow \mathcal{A}$ to interact with the environment. At a time step $t$, the agent takes action $a_t=\pi(s_t)$ according to the current state $s_t$. The environment state will transfer to next state $s_{t+1}$ based on the transition probability $\mathcal{T}$. The agent will receive the reward $r_t=R(s_t, a_t, s_{t+1})$. Then, the next round of interaction begins. The goal of this agent is to find the optimal policy $\pi^*$ that maximizes the expected return: $\pi^* = argmax_{\pi} \mathbb{E}[\sum_{t=0}^{\infty}\gamma^t r_t | \pi]$.

Q-learning \cite{watkins1992q} is a well-known RL algorithm that obtains the optimal policy by learning its $Q$ function. The $Q$ function for policy $\pi$ on a state-action pair $(s, a)$ is defined as follows.
\begin{eqnarray}
	Q(s, a) = \mathbb{E}_{\pi,\mathcal{T}}[\sum_{t=0}^{\infty}\gamma^tR(s_t, a_t, s_{t+1})|s_0=s,a_0=a]
	\label{equ:q}
\end{eqnarray}
In Q-learning, the agent learns to estimate $Q(s, a)$ by updating it based on the experience $(s_t, a_t, s_{t+1}, r_t)$. After the estimated $Q$ function converges to optimal $Q^*$, the corresponding policy $\pi(s) = argmax_aQ^*(s, a)$ at any state $s\in\mathcal{S}$ is an optimal policy.


In this work, we extend the standard RL problem with domain knowledge consisting of a finite vocabulary set $\mathcal{P}$ and a labeling function $L: S \rightarrow \mathcal{P} \cup \{\emptyset\}$. A vocabulary $p\in\mathcal{P}$ is seen as a subtask that is potentially helpful for obtaining rewards in the environment. Given $\mathcal{P}$ and the labeling function $L$, $L(s)=p$ means that $p$ is the subtask that is achieved at state $s$ and $L(s)=\emptyset$ means that no subtask is achieved at state $s$.  Note that $L(s)$ does not represent the subtasks that have been achieved prior to state $s$. 

\begin{figure}[!h]
	\centering
	\subfigure[]{\includegraphics[width=4.5cm]{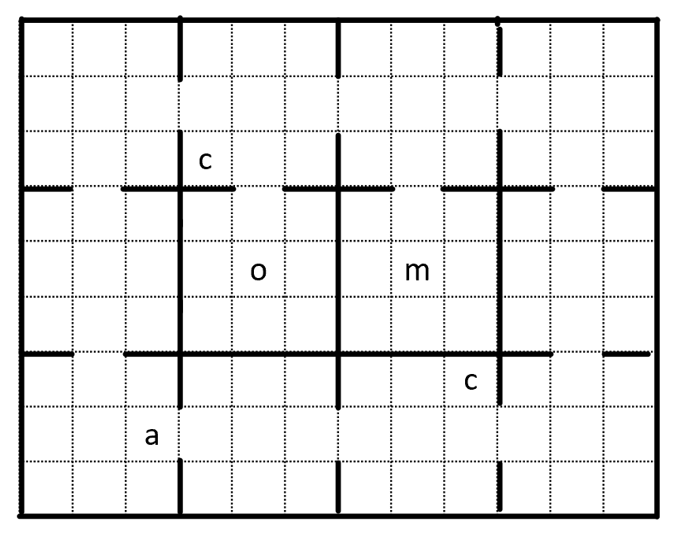}	\label{fig:office_task}}
	\subfigure[]{\includegraphics[width=3.5cm]{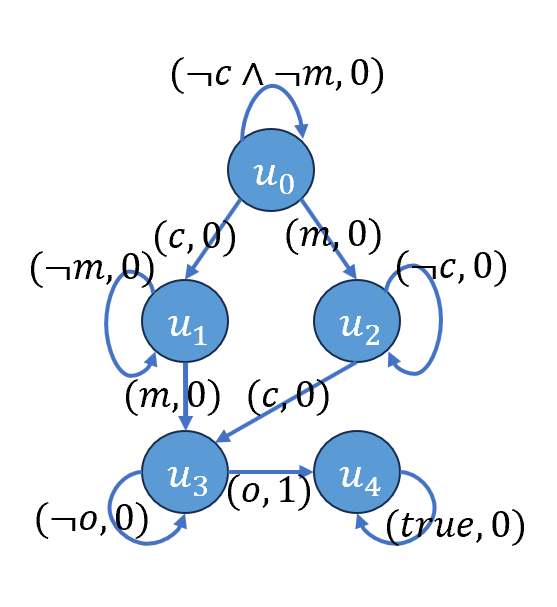} \label{fig:office_rm}}
	\caption{(a) \textit{Coffee\&mail} task on \textit{OfficeWorld} domain, where `a', `c', `m' and `o' in the figure indicates the positions of the agent, coffee, mail and office respectively. In this task, the agent is rewarded only when it arrives at the office after taking a coffee and a mail. (b) Reward machine introduced by \cite{QRM-j} to expose the reward structure of this task to RL agent.  }
	\label{fig:office}
\end{figure}

A specific example is in the \textit{OfficeWorld} domain \cite{QRM, rm2022}. As shown in Figure \ref{fig:office_task}, the vocabulary set $\{c, m, o\}$ specifies three possible subtasks in this environment. During interaction with the environment, agents can use $L$ to detect whether a subtask has been completed at $s$. For example, $L(s)=c$ means the subtask `picking up coffee' is achieved by the agent at $s$. Likewise, $L(s)=o$ means at state $s$ the agent finishes the subtask `arriving at office'. $L(s)$ could also be $\emptyset$, which means no subtask achieved at $s$.

In recent literature, the solution to such problem is to specify \cite{QRM, QRM-j} or learn \cite{DBLP:conf/aips/0005GAMNT020, DBLP:conf/aaai/HasanbeigJAMK21} a reward machine, which is defined as a tuple $<U, u_0, F, \delta_u, \delta_r>$ where $U$ is a finite set of states, $u_0$ is an initial state, $F$ is a finite set of terminal states, $\delta_u$ is the state-transition function and $\delta_r$ is the state-reward function \cite{QRM-j}. An example of a reward machine for \textit{Coffee\&mail} task is shown in \ref{fig:office_rm}. An RL agent with this reward machine starts from $u_0$ at the beginning of an episode. If at any MDP state $s$ that $L(s)=c$, then the RL agent moves to $u_1\in U$ with reward $0$. Similarly, if at $u_3\in U$ the RL agent encounters an MDP state $s$ that $L(s)=o$ then the RL agent moves to terminal state $u_4$ with reward $1$. With such a reward machine, previous methods such as \cite{QRM, DBLP:conf/aips/0005GAMNT020, DBLP:conf/aaai/HasanbeigJAMK21} learn Q-values over the cross-product $Q(s, u, a)$, which allows the agent to consider the MDP state $s$ and RM state $u$ to select the next action $a$ \cite{QRM-j}.

Different with previous works \cite{TL1,QRM,QRM-j}, in our setting, less domain knowledge about the task is required and given. The knowledge given by $\mathcal{P}$ and $L$ is limited to subtasks and not to, for example, their ordering structures. With such a minimal knowledge to the task, the agent does not know the information how the subtasks contributes to rewards before training. In our setting, this information will be learned during the training.

\section{Methodology}

In this section we present our two-level policy design and the detailed training for the policies.

\subsection{Two-level policy formalization}
\label{subsec:two-level}

When the domain knowledge can specify and structure the reward function, the RL agent can use them to improve sample efficiency \cite{QRM-j}. When the domain knowledge minimally specifies and structures the reward function, as in our case, an inertial solution is needed to learn an exact automaton model describing the reward structure and then use it to improve RL \cite{DBLP:conf/aips/0005GAMNT020, DBLP:conf/aaai/HasanbeigJAMK21}. However, learning an exact automaton model from trace data is known to be NP-complete \cite{DBLP:journals/iandc/Gold78}, which could make the algorithm computationally intractable. Besides, when the learned automaton model is inaccurate or partial to specify the reward function, the RL agent may fail to learn optimal policy. Therefore, instead of learning an exact model with domain knowledge, in this work we propose a more efficient way to integrate domain knowledge into the learning procedure of an RL algorithm.

In the framework of ALCS, a high-level policy is designed to select next subtasks to be achieved based on the environment state and the sequence of subtasks that have already achieved in the history this episode.  We denote it as $\pi_h: \mathcal{S} \times \mathcal{P}^* \rightarrow \mathcal{P}$, where $\mathcal{P}^*$ is the Kleene closure on $\mathcal{P}$. For example, in Figure \ref{fig:office} where $\mathcal{P}=\{c,m,o\}$, the Kleene closure on $\mathcal{P}$ is $\mathcal{P}^*=\{\emptyset, c, m, o, cc, cm, co, mc, mm, mo, oc, om, \\oo, ccc, ....\}$. Given a state $s$, $\pi_h$ selects the next subtask $p$ by:
\begin{eqnarray}
	p = \pi_h(s, p^*) \label{equ:pi_high}
\end{eqnarray}
where the sequence $p^*\in\mathcal{P}^*$ represents the order of subtasks have been temporally achieved in the history of a given episode. We believe that the completed subtasks and their order are key information for $\pi_h$ to select the next subtask to be achieved. Therefore, the agent will benefit from training the policy $\pi_h$ with such sequence being considered.

On the other hand, a low-level policy, denoted as $\pi_l: \mathcal{S} \times \mathcal{P} \rightarrow \mathcal{A}$, is designed to learn to achieve the selected subtask efficiently. Taking the current state and a given subtask, $\pi_l$ decides an environment action to achieve the subtask. $\pi_l$ selects actions by:
\begin{eqnarray}
	a = \pi_l(s, p) \label{equ:pi_low}
\end{eqnarray}
With the two-level policies, our agent interacts with the environment as follows. At the beginning of an episode, the agent initializes an empty sequence $p^*$ to store the achieved subtasks in the environment. At each time step $t$, the agent first employs the high-level policy to select a subtask $p_t$ to be achieved based on the current state $s_t$ and the achieved subtask sequence $p^*_t$, i.e., $p_t = \pi_h(s_t, p^*_t)$. Then, the agent uses the low-level policy to choose action $a_t = \pi_l(s_t, p_t)$ to interact with the environment and receive the reward $r_t$ the next state $s_{t+1}$. Moreover, if a subtask is achieved in $s_{t+1}$, i.e., $L(s_{t+1})\neq\emptyset$, then the achieved subtask will be appended into the sequence:
\begin{eqnarray}
	p^*_{t+1} = \begin{cases}
		p^*_{t} \oplus L(s_{t+1}) & \text{if}\quad  L(s_{t+1})\neq\emptyset \\p^*_{t}&\text{otherwise}
	\end{cases}\label{equ:q_t}
\end{eqnarray}
where $\oplus$ represents appending $L(s_{t+1})$ into the end of sequence $p^*_{t}$.

Following Q-learning, our $\pi_h$ and $\pi_l$ can select subtasks and actions by doing argmax on their corresponding $Q$ functions. We next describe the definition of their $Q$ functions and the detailed corresponding training process.

\subsection{Low-level training}
\label{subsec:low-level}

In order to define the low-level Q function, we first define the low-level reward function. The goal of low-level policy $\pi_l$ is to achieve the given subtask $p\in\mathcal{P}$. Therefore, in our design, the low-level policy does not directly maximize the expected return from the environment reward function $R$. For a certain $p\in\mathcal{P}$, $\pi_l(s, p)$ is trained with the following rewards:
\begin{eqnarray}
	R^p(s_t, a_t, s_{t+1}) = \begin{cases}
		1 & \text{if}\quad  p = L(s_{t+1}) \quad\text{and}\quad p \neq L(s_{t}) \\0&\text{otherwise}
	\end{cases}\label{equ:r}
\end{eqnarray}
According to Equation (\ref{equ:r}), it can be known that for different subtasks, $\pi_l$ is trained with different reward functions.
Given Equation (\ref{equ:r}), we define the low-level $Q$ function $Q_l(s, p, a)$ as the excepted return based on the subtask $p$: 
\begin{eqnarray}
	Q_l(s, p, a) = \mathbb{E}_{\pi_l,\mathcal{T}}[\sum_{t=0}^{\infty}\gamma^tR^p(s_t, a_t, s_{t+1})|s_0=s,a_0=a]
	\label{equ:q_l}
\end{eqnarray}

With the defined low-level $Q_l$ in Equation (\ref{equ:q_l}), $\pi_l$ can  select actions by doing argmax operation: $a_t = argmax_{a}Q_l(s_t, p, a)$ at time step $t$. Then we can train $\pi_l$ by updating $Q_l$ with experience $(s_t, a_t, s_{t+1}, r_t^p)$ using Q-learning, where $r_t^p$ is the reward value computed from Equation (\ref{equ:r}) with the current subtask $p$ to be achieved.

We observe that the reward function in Equation (\ref{equ:r}) is computed by labeling function $L$. Inspired by counterfactual experiences \cite{QRM-j}, to improve the training efficiency, we use $L$ to generate multiple experience for different subtasks to update $Q_l$, instead of using a single experience $(s_t, a_t, s_{t+1}, r^p_t)$ to update it. To do so, for a single transition $(s_t, a_t, s_{t+1})$ sampled from the environment, we can generate a set of experiences: $\{(s_t, a_t, s_{t+1}, r^p_t)\}$ for all $p\in\mathcal{P}$, where $r^p_t = R^p(s_t, a_t, s_{t+1})$ is the reward value with respect to subtask $p$ at time step $t$.

\begin{algorithm}[htb]\footnotesize
	\caption{Generating multiple experiences and updating $Q_l$.}
	\label{alg:low-level}
	\hspace*{\algorithmicindent} \raggedright\textbf{Input:} Transition $(s_t, a_t, s_{t+1})$, original $Q_{l}$, learning rate $\alpha$ \\
	\hspace*{\algorithmicindent} \raggedright\textbf{Output:} Updated $Q_{l}$
	\begin{algorithmic}[1]
		\State Initialize $experiences \leftarrow \{\}$ 
		\For{$p \in \mathcal{P}$}
		\If{$p = L(s_{t+1})$ and $p \neq L(s_{t})$}
		\State $r^p_t \leftarrow 1$, $done \leftarrow True$ 
		\Else
		\State $r^p_t \leftarrow 0$, $done \leftarrow False$ 
		\EndIf
		\State Add tuple $(s_t, a_t, s_{t+1}, r^p_t, p, done)$ into $experiences$
		\EndFor
		\For{$(s, a, s', r, p, done)$ in $experiences$}
		\If{$done$} 
		\State $y \leftarrow r$
		\Else
		\State $y \leftarrow r+\gamma\max_{a'}Q_l(s', p, a')$
		\EndIf
		\State $Q_l(s, p, a) \leftarrow (1-\alpha)Q_l(s, p, a) + \alpha \cdot y $
		\EndFor
	\end{algorithmic}
\end{algorithm}

Algorithm \ref{alg:low-level} describes the detailed practice for updating $Q_l$ with multiple experiences generated from single transition. Given an input transition $(s_t, a_t, s_{t+1})$, Algorithm \ref{alg:low-level} in Lines $2\sim7$ assigns different rewards following Equation \ref{equ:r} for all possible $p\in \mathcal{P}$ to this transition, which yields a collection of experiences. Then those experiences will be used to update $Q_l$ in Lines $10\sim17$.

After sufficiently updating for $Q_l$, given a subtask $p$, $\pi_l$ can repeatedly perform this subtask. But this is generally not enough to maximize the environment rewards. So we also need to train the $\pi_h$, which is responsible for learning to compose subtasks in any order to maximize the environment rewards.

\subsection{High-level training}
\label{subsec:high-level}

The goal of the high-level policy $\pi_h$ is to compose subtasks for the original task in MDP. So we use the environment original reward to train the $\pi_h$. The corresponding $Q$ function for $\pi_h$ is denoted as $Q_h(s, p^*, p)$, which is defined as the excepted return from the environment following $\pi_h$ and $\pi_l$ if the agent selects $p$ as a subtask given $s$ and $p^*$:
\begin{eqnarray}
	Q_h(s, p^*, p) = \mathbb{E}_{\pi_l,\pi_h,\mathcal{T}}[\sum_{t=0}^{\infty}\gamma^tR(s_t, a_t, s_{t+1})|s_0=s,a_0=\pi_l(s,p)]
	\label{equ:q_h}
\end{eqnarray}
With the high-level function defined in Equation (\ref{equ:q_h}), $\pi_h$ selects a sub-goal $p_t$ by doing the argmax operation: $p_t = argmax_{p}Q_h(s_t, p^*_t, p)$ for given $s_t$ and $p^*_t$.

In the general RL methods, $Q$ function should be updated based on the outputs from the corresponding policy. However, when this approach is applied to our high-level training, it may introduce the problem of multiple subtasks being completed under one target subtask. This can lead to decisions about key subtasks not being learned. For example, a sub-goal $p_t$ is chosen by $\pi_h(s_t, p^*_t)$. When $\pi_l(s_t, p_t)$ takes actions to achieve $p_t$, another subtask $p'$ could be completed. In this case, if $p'$ is the key subtask that brings reward in future time step, the importance of $p'$ to reward cannot be learned because $\pi_h(s_t, p^*_t)$ never choose $p'$ as its decision.

\begin{figure}[!t]
	\centering
	\includegraphics[width=0.7\linewidth]{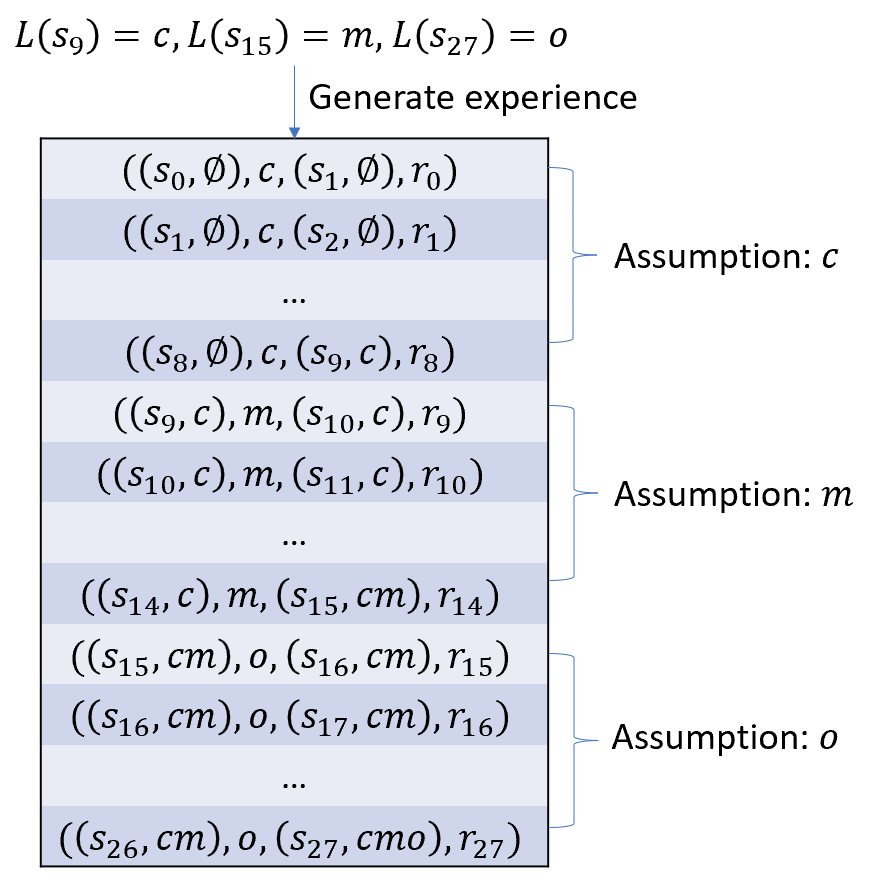}
	\caption{Following the environment and task of Figure \ref{fig:office}, here is an example about how we generate the high-level experiences in an episode for updating $Q_h$. As shown in the table, each row contains the corresponding experience data at that time step. The experience is denoted as: $((s_t, p^*_t), p_t, (s_{t+1}, p^*_{t+1}), r_t)$, where $p_t$ here is the assumed selected subtask and $r_t$ is reward from the environment. In this episode, subtask $c$, $m$, and $o$ are achieved at time step $9$, $15$, and $27$, respectively. Then, the assumed subtask $p_t$ selected by $\pi_h$ are $c$ from time step $0\sim8$, $m$ from time step $9\sim14$, and $o$ from time step $15\sim27$.}
	\label{fig:experience}
\end{figure}

To solve this problem, we propose to generate the high-level experiences based on the actual completed subtask in the environment (subtask $p'$ in the above example), instead of based on the subtask selected by the high-level policy (sub-task $p$ in the example). In other words, the actual completed task will be rewarded and used in the corresponding data $((s_t, p^*_t), p', (s_{t+1}, p^*_{t+1}), r_t)$ for the update. The basic idea is that whenever a subtask is detected to have been achieved by the labeling function, we assume that $\pi_h$ has chosen this subtask. And experiences will be generated to update $Q_h$ based on this assumption. An example for generating experiences in an episode is shown in Figure \ref{fig:experience}.

We also would like to clarify that in our design the exploration of high-level policy is stochastic. In fact, in our two-level policy design, as long as action selection in low-level is stochastic, high-level policy will also perform stochastic exploration. Because the stochastic actions cause the substasks being achieved stochastically, the assumed high-level choice will be stochastic as well.

\begin{algorithm}[h]\footnotesize
	\caption{Automatically Learning to Compose Subtasks.}
	\label{alg:framework}
	\hspace*{\algorithmicindent} \raggedright\textbf{Input:} Total episode $M$, learning rate $\beta$, exploration rate $\epsilon$\\
	\hspace*{\algorithmicindent} \raggedright\textbf{Output:} $Q_{l}, Q_{h}$
	\begin{algorithmic}[1]
		\For{$episode = 1 \to M$}
		\State Reset $Env$ and get $s_0$, $t \leftarrow 0$
		\State $experience\_h \leftarrow \{\}$, $e\_{temp} \leftarrow \{\}$
		\State $p^*_t \leftarrow []$  
		\While{not terminal}
		\State Obtain subtask $p_t = argmax_{p}Q_h(s_t,p^*_t,p)$
		\If{$rand()>\epsilon$}
		\State Take the action $a_t = argmax_{a}Q_l(s_t,p_t,a)$
		\Else
		\State Take a random action as $a_t$
		\EndIf
		\State Execute $a_t$ and get $s_{t+1}$ and $r_t$ from $Env$
		\State Update $Q_{l}$ using Algorithm \ref{alg:low-level} with $(s_t, a_t, s_{t+1})$  
		\If{$L(s_{t+1}) = \emptyset$}
		\State $e\leftarrow((s_t, p^*_t), p_t, (s_{t+1}, p^*_{t}), r_t)$ 
		\State Add experience tuple $e$ into $e\_{temp}$
		\State $p^*_{t+1} \leftarrow p^*_{t}$
		\Else
		\State $p_{act} \leftarrow L(s_{t+1})$
		\State $p^*_{t+1} \leftarrow p^*_t \oplus p_{act}$
		\State $e\leftarrow((s_t, p^*_t), p_t, (s_{t+1}, p^*_{t+1}), r_t)$ 
		\State Add experience tuple $e$ into $e\_{temp}$
		\For{$((s, p^*), p, (s', p^{*'}), r)$ in $e\_{temp}$}
		\State $e\_assumed\leftarrow((s, p^*), p_{act}, (s', p^{*'}), r)$
		\State Add $e\_assumed$ into $experience\_h$
		\EndFor  
		\State $e\_{temp} \leftarrow \{\}$
		\EndIf                      
		\If{terminal}
		\For{$((s, p^*), p, (s', p^{*'}), r)$ in $experiences\_h$} 
		\State $y \leftarrow r+\gamma\max_{p'}Q_h(s', p^{*'}, p')$
		\State $Q_h(s, p^*, p) \leftarrow (1-\beta)Q_h(s, p^*, p) + \beta \cdot y $
		\EndFor        
		\State $experiences\_h \leftarrow \{\}$
		\EndIf
		\State $t \leftarrow t+1$
		\EndWhile
		\EndFor
	\end{algorithmic}
\end{algorithm}

\subsection{Overall algorithm}

Algorithm \ref{alg:framework} shows the overall training procedure of our method. When an episode begins, we initialize some variables in Lines $2\sim4$, where $experience\_h$ is the experiences for this episode to update $Q_h$, $e\_temp$ is a temporary experience set, and $p^*_t$ stores the sequence of achieved subtasks at the time step $t$. After the initialization for an episode, the algorithm will repeat the process in Line $6\sim36$ until the episode terminates. In each episode, the agent first choose a subtask $p_t$ in Line $6$. Then an action $a_t$ is selected following the $\epsilon-$greedy exploration in Line $7\sim11$, where $rand()$ represents a function for sampling a variable between 0 and 1 uniformly. After executing $a_t$, the $Q_l$ is updated using Algorithm \ref{alg:low-level} with respect to the current transition in Line $13$. Then, high-level experiences are generated in Line $14\sim 28$. The experiences $e$ in line $15$ and Line $21$ are differentiated according to whether there is a subtask being achieved at current time step. If a subtask is achieved (i.e., $L(s_{t+1}) \neq \emptyset$), this achieved subtask, denoted as $p_{act}$ in Line $19$, will be appended into $p^*_t$ in Line $20$. Besides, $p_{act}$ will be treated as the assumed subtask selected by $\pi_h$. Based on $p_{act}$ the high-level experiences for updating are generated in Line $23\sim 26$. Finally, $Q_h$ will be updated for each experience in the episode in Line $30\sim 33$.

\section{Interpretability}
\label{sec:inter}

How to make the RL algorithms interpretable has widely raised concerns recently for ethical and legal considerations in application \cite{exRL2020, exRL2021, exRL2022}. In this section, based on the given minimal domain knowledge, we interpret the behavior of agents trained by ALCS.

\begin{figure}[!h]
	\centering
	\includegraphics[width=0.75\linewidth]{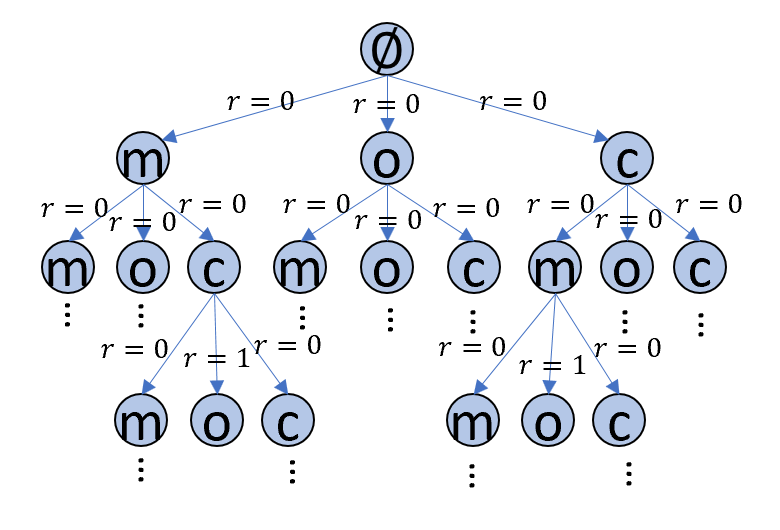}
	\caption{An example tree to record sequences of subtasks in the environment of Figure \ref{fig:office}. This tree is generated during the training process. Except for the root, nodes in this tree refer to the subtasks specified in this environment. The edges present the transferring from one subtask to another subtask (when the parent subtask has been achieved). The reward values on the edges are sampled from the environment based on whether the main task is achieved. Following the task description in Figure \ref{fig:office}, `$r=1$' means the task is finished, i.e., the agent arrives at the office after taking both a coffee and a mail. For infinite MDP where the episode step can go to infinity, the tree could be infinite. For finite MDP where the episode step is limited, the tree is finite because the depth of the tree will be limited by the episode steps and the width will be limited by $|\mathcal{P}|$.}
	\label{fig:tree}
\end{figure}

We can use a tree sturcture to record all the sequence of subtasks that have been achieved in the training process of ALCS, which can be used to interpret the behavior of an agent.
This tree begins with a root node `$\emptyset$'. Each descendant node of the root is a subtask $p\in \mathcal{P}$. The nodes in the path from the root to a descendant can form a sequence $p^*\in\mathcal{P}^*$. In the beginning of an episode, the agent always starts from the root node `$\emptyset$', which means no subtask has been achieved in the history of this episode. Then, if a subtask $p$ is achieved, a new child node with respect to $p$ is added in the tree. Besides, the reward from environment when subtask $p$ is achieved will be recorded on the edge to the corresponding child node. Given an achieved sequence of subtask $p^*_t$ at time step $t$, a node associated with the current state can be uniquely identified in the tree. We call this node as a current node at $t$, denoted as $N(p^*_t)$. At the start of the episode, the agent's current node $N(p^*_t)$ is reset to the root node, i.e. $p^*_t$ is reset to $\emptyset$. An example for this is shown in Figure \ref{fig:tree}.

With this tree, the RL agent can interpret its behavior based on the given domain knowledge during execution. The interpretation consists of the following three parts.

\textbf{What has already happend.} At a certain time step during execution, according to the current $p^*_t$, we can identify the current node of the trained RL agent. Because $p^*_t$ includes the historically achieved subtasks, which can be seen as an abstraction for the episode history based on domain knowledge. Therefore, $p^*_t$ can be used as an interpretation of what has happened.

\textbf{Current best subtask.} Based on the current selected $p_t$, the agent can interpret its current action $a_t$ in terms of which subtask this action is trying to complete. For example, if $c$ is given from $\pi_h$ in the example task of Figure \ref{fig:office}, then the action from $\pi_l$ can be interpreted as an action trying to complete $c$.

\textbf{Planning for the future.} After obtaining $p_t$, the agent can find the child node of the current node. By performing a breadth-first search (BFS) on the subtree with this child node as root, the agent can find a descendant node that can obtain the reward based on the previous record. By providing the nodes on this path, the agent can interpret an estimated future planning for the current selected subtask. Taking Figure \ref{fig:tree} as an example, suppose current $p^*_t = \emptyset$ and $p_t = c$, then the estimated future plan is $m$ and $o$, considering that the BFS algorithm will find the satisfied descendant node with minimum depth first and return it. If the BFS algorithm cannot find a satisfied descendant node in limited depth, then we say there is no interpretation to future plan for the current decision.

The advantage of this is that the interpretation takes into account both the high-level information constructed by domain knowledge (i.e., the tree) and the environment state information through $p_t=\pi(s_t, p^*_t)$. When the environment state changes, both the behavior and its interpretation could change, which makes agent more transparent on its decision making procedure. We will further demonstrate this benefit in Subsection \ref{subsec:int}.

\section{Experiments}

This section presents the experiment settings and results\footnote{The code is released (anonymously) at: https://github.com/anonymousanonymousd/ALCS.}.

\subsection{Experiment settings}

\setlength{\abovecaptionskip}{0.cm}
\begin{figure*}[!h]
	\centering
	\subfigure[Coffee]{\includegraphics[width=4cm]{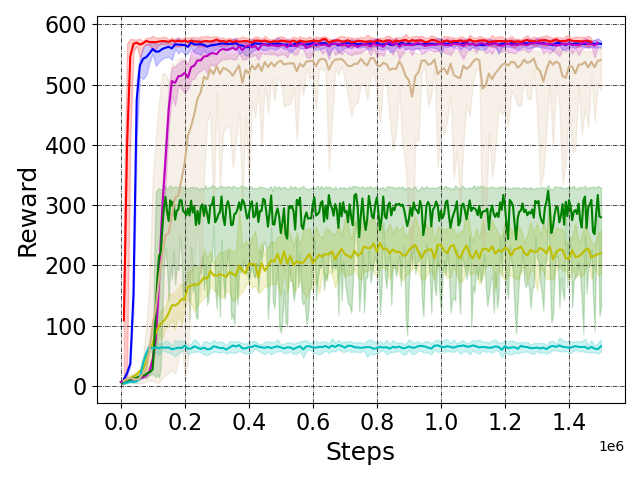}}
	\subfigure[Coffee and Mail]{\includegraphics[width=4cm]{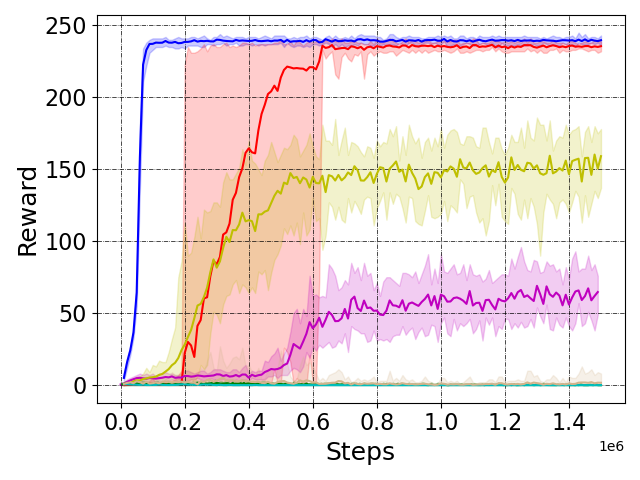}}
	\subfigure[Collecting]{\includegraphics[width=4cm]{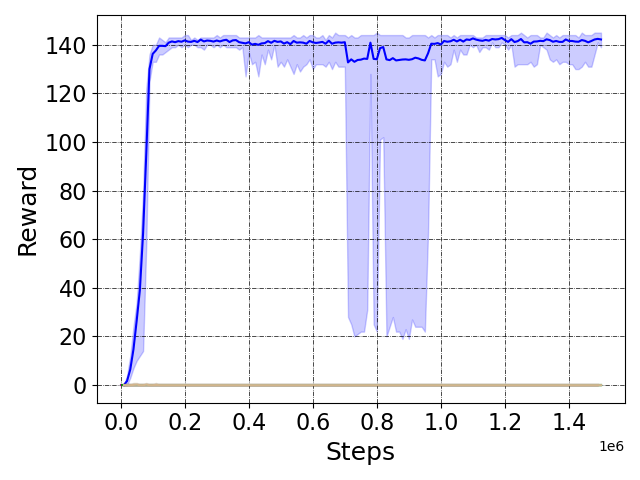}}
	\subfigure[Bonus]{\includegraphics[width=4cm]{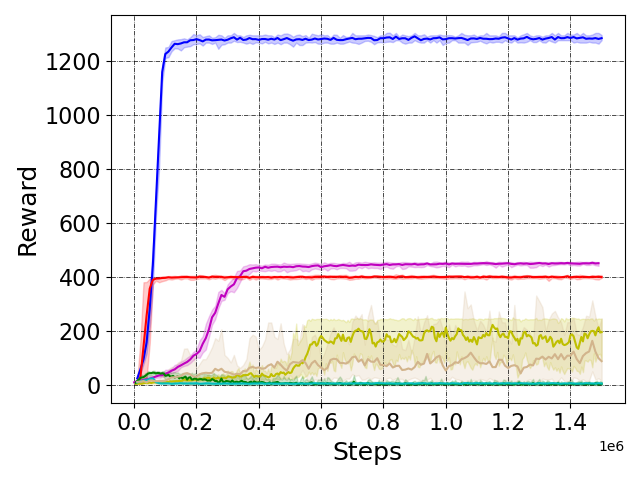}}
	
	\subfigure[Plant]{\includegraphics[width=4cm]{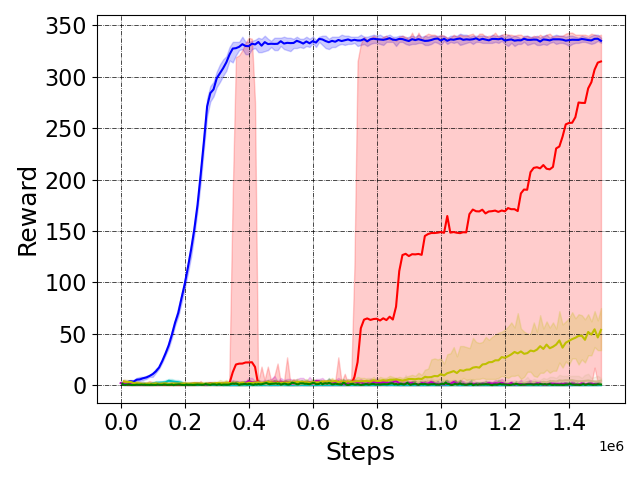}}
	\subfigure[Bridge]{\includegraphics[width=4cm]{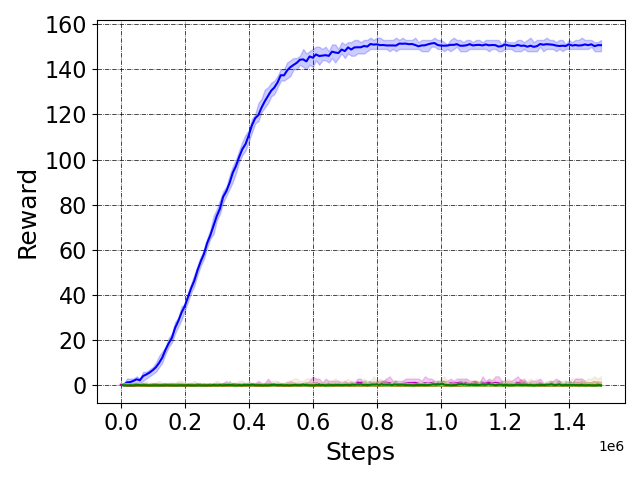}}
	\subfigure[Bed]{\includegraphics[width=4cm]{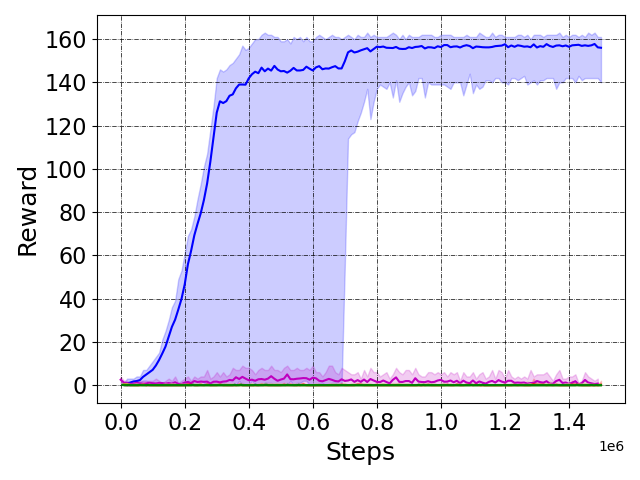}}
	\subfigure[Gem]{\includegraphics[width=4cm]{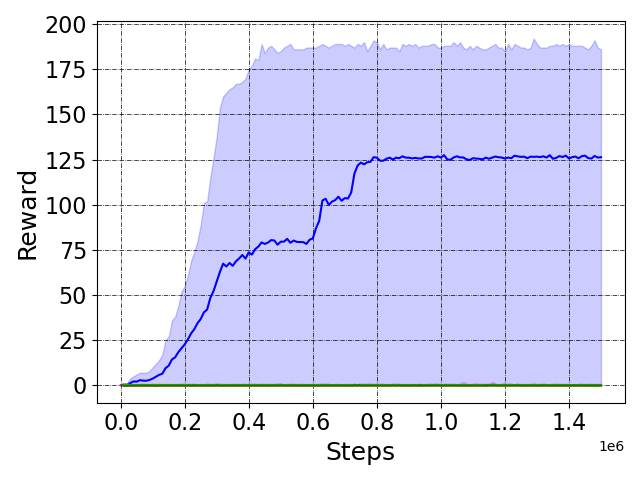}}
	
	\subfigure{\includegraphics[width=15cm]{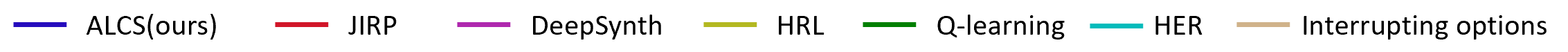}}
	\caption{Learning curves of various RL algorithms on 8 environments from \textit{OfficeWord} and \textit{MineCraft} domains.}
	\label{fig:compare}
	\Description{Comparison with baselines.}
\end{figure*}

We evaluate our method in 8 different sparse-reward envrionments from the two commonly used domains, \textit{OfficeWord} and \textit{MineCraft} \cite{QRM, QRM-j, DBLP:conf/aips/0005GAMNT020}.  We introduce the characteristics of these environments and the reasons for choosing them as follows.

\textbf{Coffee.} Go to office after taking a coffee. The agent recieves 1 reward after achieving this task. 

\textbf{Coffee$\&$Mail.} Go to office after taking both a coffee and a mail. The agent recieves 1 reward after achieving this task. The order of taking the coffee and mail does not matter.

\textbf{Collecting.} Go to office after collecting four packages A, B, C and D in the corners. The agent recieves 1 reward after achieving this task. The order of taking those packages does not matter.

\textbf{Bonus.} Same task with the Collecting environment. When agent arrive at office, it will receive 1 reward for each package, and 5 bonus reward if all packages are collected.

\textbf{Plant.} Get wood, use toolshed. The agent recieves 1 reward after achieving this task. 

\textbf{Bridge.} Get iron, get wood, use factory. The agent recieves 1 reward after achieving this task. The iron and wood can be gotten in any order.

\textbf{Bed.} Get wood, use toolshed, get grass, use workbench. The agent recieves 1 reward after achieving this task. The grass can be gotten at any time before using the workbench.

\textbf{Gem.} Get wood, use workbench, get iron, use toolshed, use axe. The agent recieves 1 reward after achieving this task. The iron can be gotten at any time before using the toolshed.
 
In the above environments, the former four environments are from the \textit{OfficeWord} domain, while the latter four are from \textit{MineCraft}. These environments have increasing task complexity in their respective domains, which is helpful to show how the performance of methods changes as the task difficulty increases. Moreover, \textit{MineCraft} domain has a larger state space than \textit{OfficeWord}. Comparing the performance of methods in the two domains can help to demonstrate the impact of the increased state space on the performance of different methods. 

In addition, Bonus is a special environment. In this environment, the agent can get 1 reward for each package, but prefers to collect all the packages. Note that the goal in the Bonus environment is too sparse for providing rewards. Thus, in this environment, there is also environment reward for completing individual subtasks (collecting one or more packages), which helps the baseline methods learn the skill of collecting packages. But at the same time those reward can lead to  the agent misalignment policy from learning to collect all packages to learning to collect individual packages. Therefore in this environment, it is challenging for the RL agent to utilize the subtask reward without such policy misalignment.

We compare the following five different methods.

\textbf{ALCS (ours)}. We implement a prototype of ALCS, which uses the Q-learning algorithm \cite{watkins1992q} to learn both the high-level and low-level policies. The learning rates for both $Q_l$ and $Q_h$ are $0.1$. In the training, the rate for $\epsilon-$greedy exploration is $20\%$ to provide a stochastic exploration strategy. 

\textbf{JIRP}. JIRP \cite{DBLP:conf/aips/0005GAMNT020} is a method that infers a reward machine using the $libalf$ library \cite{bollig2010libalf} as the RL agent interacts with the environment. We chose JIRP as a baseline because it has a similar problem setting to our work, i.e., how to automatically learn to structure the unknown reward function through a minimal domain knowledge.

\textbf{DeepSynth}. DeepSynth \cite{DBLP:conf/aaai/HasanbeigJAMK21} is an algorithm that infers a automaton model to guide RL agent to learn policy. This method uses automata synthesis \cite{jeppu2020learning} as an approach to learn a high-level model. Likewise, DeepSynth has a similar problem setting to our work.

\textbf{HRL}. HRL \cite{kulkarni2016hierarchical} learns hierarchical policies in the environment. Following previous implementation \cite{DBLP:conf/aips/0005GAMNT020}, the option for high-level policy in HRL is defined as subtask $p\in\mathcal{P}$. The termination condition of an option is whenever the corresponding $p$ is achieved. 

\textbf{Interrupting options}. Interrupting options \cite{sutton1999between} is a hierarchical RL method in which the high-level policy can interrupt low-level policies and change the current subtask to be completed. We chose HRL and Interrupting options as our baselines because their policies have similar structures to our method. 

\textbf{HER}. HER \cite{her} is a well-known goal-conditioned RL method that trains policies to achieve different goals. The goal in HER is defined as the subtask $p\in\mathcal{P}$. Each goal-conditioned policy is trained with the same reward function as Equation (\ref{equ:r}). We choose HER as one of our baselines because the training of HER goal-conditioned policies is similar with our low-level policy training. 

\textbf{Q learning}. Q-learning \cite{watkins1992q} is a well-known general RL algorithm. We use the implementation version from \cite{QRM-j}.

Notably, JIRP and DeepSynth are two state-of-the-art methods in \textit{OfficeWord} and \textit{MineCraft} domains. Generally speaking, JIRP learns better when there is only binary reward signal in the end of episode, and DeepSynth is more adaptable in non-binary cases. Except the Q-Learning which does not require subtask, the other five baselines use equivalent domain knowledge with the proposed ALCS method. They all have access to a given set of subtasks, as well as a label function that can return to agent the identity of the subtask based on the environmental state when the subtask is completed for the first time in an episode.


\subsection{Comparison}

\setlength{\abovecaptionskip}{0.cm}
\begin{figure*}[!h]
	\centering
	\subfigure[Coffee]{\includegraphics[width=4cm]{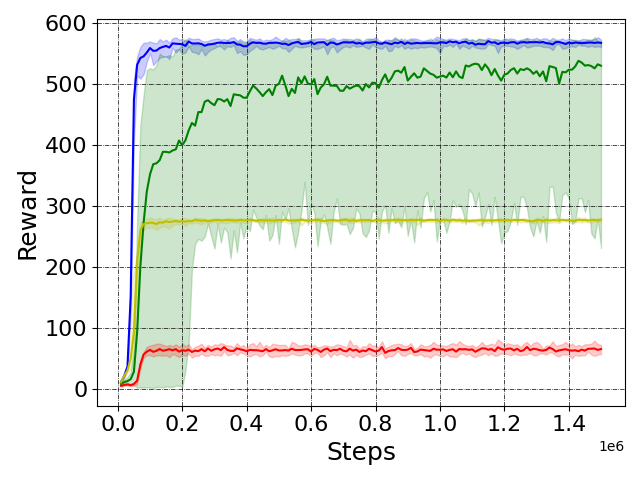}}
	\subfigure[Coffee and Mail]{\includegraphics[width=4cm]{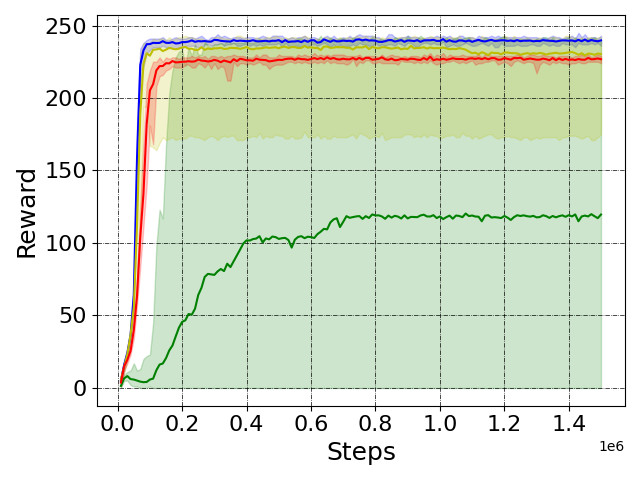}}
	\subfigure[Collecting]{\includegraphics[width=4cm]{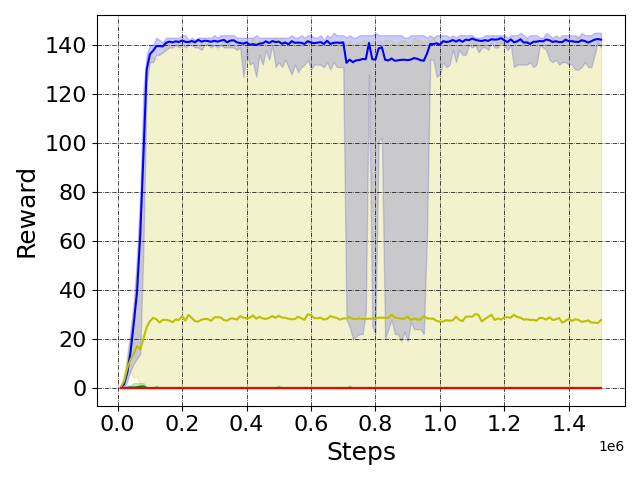}}
	\subfigure[Bonus]{\includegraphics[width=4cm]{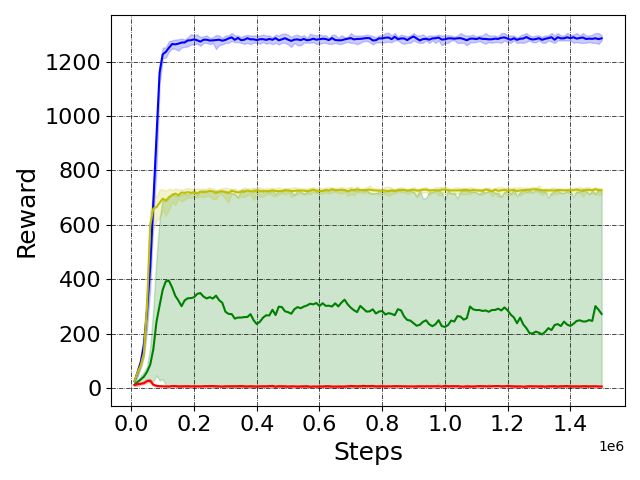}}
	
	\subfigure{\includegraphics[width=10cm]{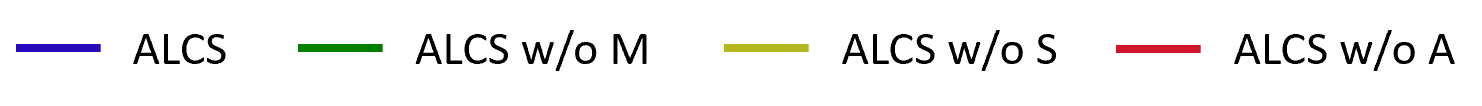}}
	\caption{Learning curves for ablation experiment of ALCS on 4 environments from \textit{OfficeWord}.}
	\label{fig:ab}
	\Description{Comparison with baselines.}
\end{figure*}

We first compare our method with baselines on 8 environments from \textit{OfficeWord} and \textit{MineCraft} domains to validate the superiority of ALCS. The results are shown in Figure \ref{fig:compare}. The error bounds (i.e., shadow shapes) indicate the upper and lower bounds of the performance with 20 runs excluding $2$ best and $2$ worst results.

The results show that our method significantly outperforms the baseline methods in all environments except Coffee where the reward structure is simple and easy to explore. In such an environment, learning an exact model with JIRP in Coffee environments to specify the reward structure is a better solution. However, when the reward structure of the task becomes complex, the sample efficiency of our method can significantly outperform methods that learn models to specify the reward structure. 

Another conclusion can be drawn by comparing those methods between \textit{OfficeWord} and \textit{MineCraft} domains. It is well known that the sample efficiency of the RL algorithm can decrease on environments with large state space. Comparing the performance of those RL algorithms in \textit{OfficeWord} domain, their performance in the \textit{MineCraft} are reduced because of the larger state space. In contrast, our approach is less affected by increased state space compared to the baselines approaches. Nevertheless, by comparing the performance of ALCS across Bridge, Bed and Gem environments, it can be seen that the performance variance of ALCS grows with the increasing task complexity. This suggests that there is room to improve the performance of ALCS in complex tasks with large state spaces.

Furthermore, comparing with Collecting environment where all baselines totally fails on policy learning, these baselines can learn some policies for subtasks when rewards is shaped for completing the collection of one or more packages. However, these shaped rewards can misalign the learned policy from collecting all packages to collecting individual packages for baselines. In contrast, our two-layer policy learning can overcome such policy misalignment.

\subsection{Ablation study}

In order to explore the contribution that each proposed technique to the whole algorithm, we compare the ALCS with its ablative variants on the \textit{OfficeWord} domain. The results are shown in Figure \ref{fig:ab}, where `ALCS w/o M' stands for ALCS algorithm without multiple experiences generating for updating $Q_l$ (as proposed in Subsection \ref{subsec:low-level}), `ALCS w/o S' represents ALCS algorithm without sequence of completed subtask for $\pi_h$ (as presented in Subsection \ref{subsec:two-level}), and `ALCS w/o A' indicates ALCS algorithm without the assumed choice of subtask for training $\pi_h$ (as presented in Subsection \ref{subsec:high-level}).

As shown in Figure \ref{fig:ab}, in case of ALCS w/o M, when not generating multiple experiences for $Q_l$, the performance of the algorithm drops in all environments due to insufficient learning of the low-level policy, and in Collecting environment, ALCS w/o M cannot even learn. In ALCS w/o S, when $\pi_h$ learns without a sequence of completed subtask, it cannot select the optimal subtask to be achieved due to the inability to obtain sufficient information on the completed subtasks. In case of ALCS w/o A, we see that the assumed choice of the subtask is the key to high-level $\pi_h$ to efficiently learn to compose subtasks. In addition, in the coffee environment, the subtask on picking coffee has a high probability of being completed while executing subtask on arriving at office. In this case, the selection of subtask for picking a coffee will not be learned by the $\pi_h$. Therefore, even if the tasks on coffee environment are not complex, without assumed choice of subtask, it is difficult for ALCS to learn good policies.

\subsection{Interpretability}
\label{subsec:int}

In this subsection, we demonstrate the interpretability of our learned policy in the Coffee and Mail environment.

\begin{figure}[!h]
	\centering
	\subfigure{\includegraphics[width=7cm]{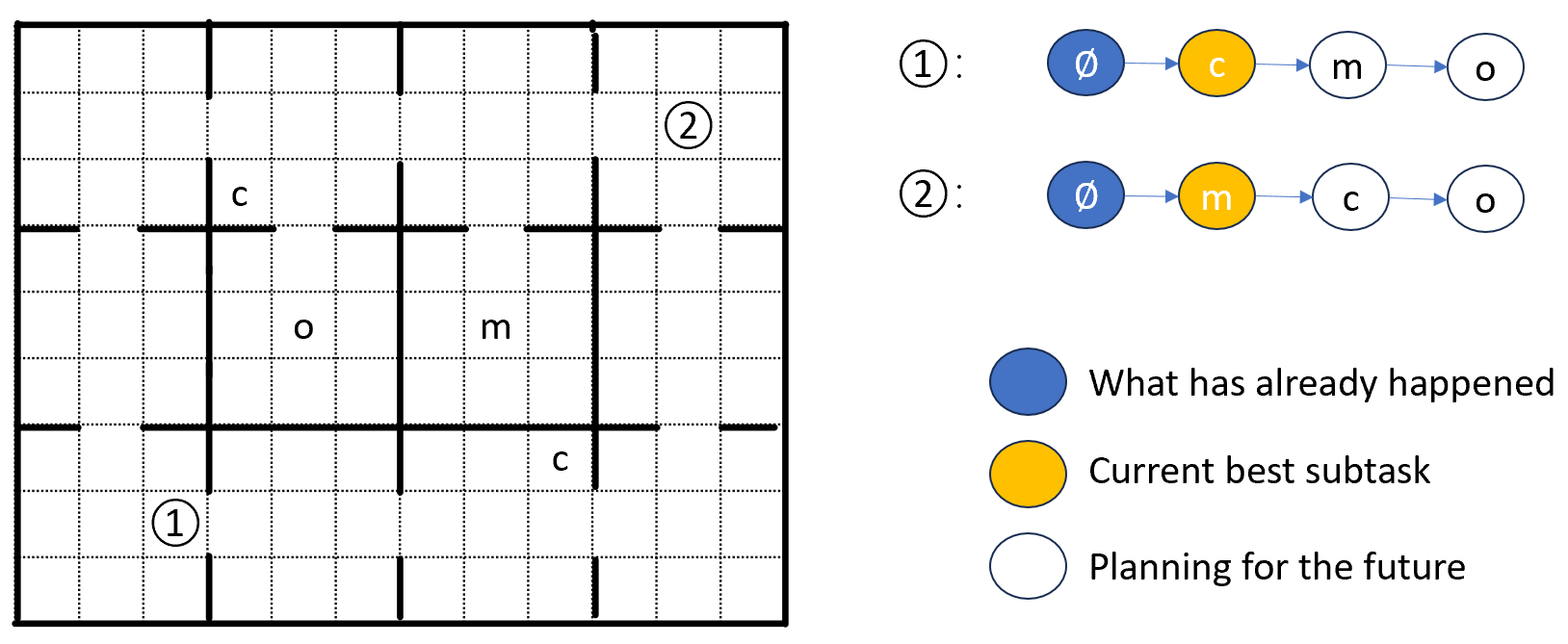}}
	\caption{Interpretation experiment on Coffee and Mail environment. The three colors on the right of the figure represent the three corresponding parts of interpretation introduced in Section \ref{sec:inter}.}
	\label{fig:inter}
	\Description{Comparison with baselines.}
\end{figure}

In this task, there are two possible sequences of subtasks to bring reward: $c\rightarrow m \rightarrow o$ and $m\rightarrow c \rightarrow o$. However, the optimal sequence of subtasks is different when the agent starts from different positions. As shown in Figure \ref{fig:inter}, starting from position \textcircled{1}, the agent takes fewer steps with the sequence $c\rightarrow m \rightarrow o$. At position \textcircled{2} $m\rightarrow c \rightarrow o$ is a better sequence. The previous approaches, by using an automaton model to specify the reward function, allows two possible sequences to be represented in a single model \cite{QRM-j, DBLP:conf/aips/0005GAMNT020}. Nevertheless, their models cannot explain that following which sequence of subtasks the agent will take fewer steps. For example, in the left of Figure \ref{fig:inter}, when starting with different positions \textcircled{1} and \textcircled{2}, the automaton model will fail to explain the agent will follow which sequence of subtasks for finishing the task with fewer steps.

The high-level policy that we are learning about can address this issue to some extent, because our high-level policy can output an exact subtask to be achieved based on the environment state. As shown in Figure \ref{fig:inter}, when the agent starts in different positions, with different state being input into $\pi_h$, $\pi_h$ will select different `current best subtask' to achieve based on Equation (\ref{equ:pi_high}). As shown in right of Figure \ref{fig:inter}, `c' is selected at \textcircled{1} and `m' is selected at \textcircled{2}. Then, according to the recording tree in Section \ref{sec:inter}, it provides a future planning after achieving the selected subtask. These constitute a complete interpretation on how to compose subtasks exactly to finish the task with fewer steps.


\section{Conclusion and future work}

In this paper, we propose ALCS, an RL algorithm to automatically structure the reward function to improve the sample efficiency in sparse-reward tasks. ALCS employs a two-level policy to learn composing subtasks and to achieve them in best order for maximizing the excepted return from environments. Besides, three optimization methods are designed for this two-level policy learning framework, which includes providing information of completed subtask sequence for better high-level decision, generating multiple experience for sample efficiency, and training high-level policy with assumed selection based on the actually achieved subtasks. We show that in a variety of sparse-reward environments, ALCS significantly outperforms the state-of-the-art methods in the environments with high task difficulty. In future work, we want to explore the convergence of the proposed algorithm and to continue our efforts to automatically learn to structure the reward function. In this paper, the subtasks is given by domain knowledge that specify subtasks. In future work, we will extend our method with partial information about the specified subtasks.

\begin{acks}
If you wish to include any acknowledgments in your paper (e.g., to 
people or funding agencies), please do so using the `\texttt{acks}' 
environment. Note that the text of your acknowledgments will be omitted
if you compile your document with the `\texttt{anonymous}' option.
\end{acks}




\bibliographystyle{ACM-Reference-Format} 
\bibliography{sample}


\end{document}